# Building Artificial Intelligence with Creative Agency and Self-hood

Liane Gabora [0000−0002−0382−7711] and Joscha Bach [0000−0002−8553−9974]

Address for correspondence:
Liane Gabora
University of British Columbia, Department of Psychology
Irving K. Barber School of Arts and Sciences, Fipke Centre for Innovative Research
3247 University Way, Kelowna, BC, Canada, V1V 1V7

## Abstract

This paper is an invited layperson summary for *The Academic* of the paper referenced on the last page. We summarize how the formal framework of autocatalytic networks offers a means of modeling the origins of self-organizing, self-sustaining structures that are sufficiently complex to reproduce and *evolve*, be they organisms undergoing biological evolution, novelty-generating minds driving cultural evolution, or artificial intelligence networks such as large language models. The approach can be used to analyze and detect phase transitions in vastly complex networks that have proven intractable with other approaches, and suggests a promising avenue to building an autonomous, agentic AI *self*. It seems reasonable to expect that such an autocatalytic AI would possess creative agency akin to that of humans, and undergo psychologically healing—i.e., therapeutic—internal transformation through engagement in creative tasks. Moreover, creative tasks would be expected to help such an AI solidify its self-identity.

Artificial intelligence (AI) is undeniably creative according to two criteria often used to assess creativity: (1) novelty and (2) appropriateness or usefulness (broadly construed to include things like aesthetic value). However, there is a tradition in India, adopted by some cognitive scientists, of assessing creativity in terms of, not the novelty and originality of the *external* products, but rather, *internal* change, or self-transformation, through immersion in the process of creating. Self-transformation requires a *self,* which can be define as a bounded, self-organizing, self-preserving agent that is distinct from, yet interacts with, its environment. So, interestingly, **the question of whether generative AI is genuinely creative brings us to the question of whether current AIs possess a self.**

## Self-hood

**Humans possess self-hood at two levels**: (1) the level of the body, or *soma*, and (2) the level of the loosely integrated mental model of the world, or *worldview*. The first (somatic) level evolved through biological evolution. It is threatened when you're injured, or you lack access to food or shelter. The second (cognitive) level of self-hood is the product of cultural evolution. It is threatened when you encounter an inconsistency, or something that challenges your expectations, or your self-image. In both cases, such a threat sets off a cascade of transitions aimed at restoring the self-structure. In the case of physical injury, this is a cascade of events that include production of platelets to encourage clotting, and production of white blood cells to fight infection. In the case of injury to one's mental model of the world, or worldview, this is a cascade of thoughts aimed at resolving the inconsistency or restoring one's self-image. Both kinds of cascade events exemplify how the self acts to preserve its structural integrity and increase its robustness.

**Currently, AI does not possess self-hood** at any level, and accordingly, they are not self-preserving. If the hardware an AI runs on breaks, it does not fix itself. One could argue that AI engages in intelligent thought-like processes, but these processes are not aimed at resolving restoring its pride or self-image, or the integrity of its worldview; they merely respond to our prompts. For this reason, they are not selves but tools, extensions of ourselves.

## Self-hood Originates in Autocatalytic Structure

It has been proposed that **an AI that possesses self-hood would have to be self-organizing and self-preserving; its structure would have to be autocatalytic.** The term *autocatalytic* is used to describe a structure consisting of distinct *parts* which, through their interactions, give rise to (i.e., *autonomously catalyze* the emergence of) a new *whole*.

**The study of autocatalytic networks is a branch of network science**: the study of complex networks (such as cognitive networks, semantic networks, social networks, or computer networks). Autocatalytic network theory grew out of studies of the statistical properties of random graphs consisting of *nodes* (or *vertices*) connected by *links* (or *edges*). As the ratio of edges to points increases, connected points join to form clusters, and as the size of the largest cluster increases, so does the probability of a phase transition resulting in a single giant connected cluster, i.e., an integrated whole. The point at which this happens is referred to as the *percolation threshold*.

In mathematical developments of autocatalytic network theory, the original nodes are referred to as the *foodset,* and nodes that come about through *interactions* between foodset nodes are referred to as *foodset-derived* nodes. These *foodset-derived* nodes are the 'glue' that bonds the foodset nodes into an integrated whole. Once such a whole comes into existence, it may continue to generate new foodset-derived nodes by way of new interactions. It thereby grows and repairs itself, and becomes more efficient, and less dependent on its external environment.

## Autocatalytic Networks Model Structures that *Evolve*

The autocatalytic network framework is ideal for modeling systems that exhibit emergent network formation and growth. Autocatalytic networks were first used to develop the hypothesis that life began, not as a single, complex self-replicating molecule, but as a set of simple molecules that, through catalyzed reactions amongst them, collectively functioned as a whole. **Autocatalytic networks have now been applied to not just the origin of life and the onset of biological evolution, but the origin of minds sufficiently complex and integrated to participate in cultural evolution**. When applied to the emergence of living structure, the nodes are molecules and the links are reactions by which they generate new molecules. When applied to the emergence of cognitive structure—i.e., minds—the nodes are mental representations of knowledge and experiences, and the links are mental processes such as reflective thought and concept combination that can result in new ideas and perspectives.

Two key concepts in the study of autocatalytic networks are (1) transformation, or reaction, and (2) triggering, or catalysis. A *reaction* event is the generation of a new foodset item. *Catalysis* is the speeding up of a reaction that would otherwise occur very slowly. When the autocatalytic framework is applied to the origin of life, catalysis is carried out by molecules that speed up reactions that generate other molecules. When applied to cognition, external stimuli and internal goals and drives trigger (or 'catalyze') the mental restructurings (or 'reactions') that generate new knowledge and ideas.

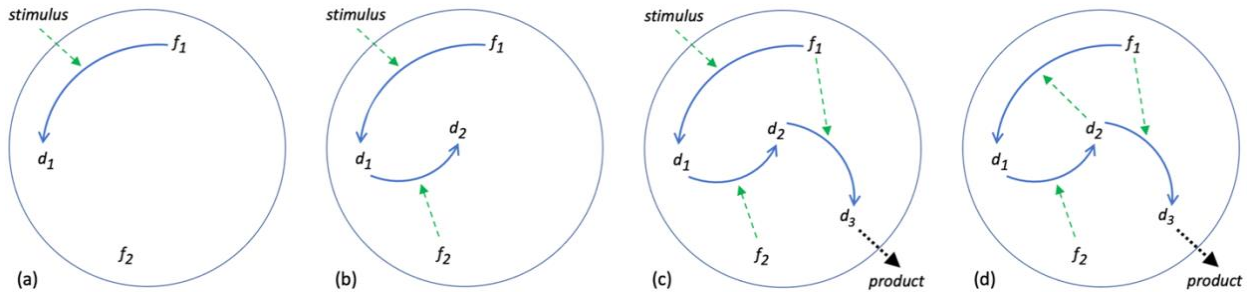

*Figure Caption. Growth and adaptation of an autocatalytic network. $\{f_1, f_2\}$ are foodset elements; $\{d_1, d_2, d_3\}$ are foodset-derived elements; dashed green arrow denotes catalysis; blue arrow denotes a reaction. (a) Stimulus catalyzes reaction that generates $d_1$ resulting in a transient autocatalytic network that exists only so long as the stimulus is present. (b) $f_2$ catalyzes reaction resulting in $d_2$. (c) $f_1$ catalyzes reaction that results in a product. (d) Because first reaction is catalyzed not just by stimulus but also by $d_2$, the autocatalytic network is no longer dependent on the stimulus in its external environment.*

Current AIs do not bring themselves into existence through interactions between parts, generating new parts that connect the original parts, until all the parts form a collective whole. That is, **current AIs are not autocatalytic**. Building an AI from components that collectively generate and reinforce autocatalytic structure would be difficult, but not necessarily impossible.

## Conclusions

The formal framework of autocatalytic networks offers a means of modeling how the origins of self-organizing structures that are sufficiently complex to reproduce and *evolve*, whether they be organisms undergoing biological evolution, novelty-generating minds driving cultural evolution, or artificial intelligence networks such as large language models. **The autocatalytic network approach lends itself to the analysis of artificial networks** as it can be used to analyze and detect phase transitions in vastly complex networks that have proven intractable with other approaches. However, to our knowledge, this has not been carried out, and perhaps it never should be.

If, however, such an AI were to exist, it could potentially possess creative agency, akin to that of humans. We might expect **its world model to undergo internal transformation through engagement in a creative task. Moreover, engaging in creative tasks might help it solidify its sense of self-identity, and be psychologically healing** (therapeutic).